\documentclass{article}

\usepackage{microtype}
\usepackage{graphicx}
\usepackage{subcaption}
\usepackage{booktabs} % for professional tables

\usepackage{hyperref}
\usepackage[accepted]{icml2026}

\usepackage{amsmath}
\usepackage{amssymb}
\usepackage{mathtools}
\usepackage{amsthm}

\usepackage{upgreek} % Added
\usepackage{xurl}

\usepackage[capitalize,noabbrev]{cleveref}

\theoremstyle{plain}

\theoremstyle{definition}

\theoremstyle{remark}

\usepackage[textsize=tiny]{todonotes}

\icmltitlerunning{Neuro-Symbolic ODE Discovery with Latent Grammar Flow}

\begin{document}

\twocolumn[
  \icmltitle{Neuro-Symbolic ODE Discovery with Latent Grammar Flow}



  \icmlsetsymbol{equal}{*}

  \begin{icmlauthorlist}
    \icmlauthor{Karin Yu}{ibk,aic}
    \icmlauthor{Eleni Chatzi}{ibk}
    \icmlauthor{Georgios Kissas}{sdsc}
  \end{icmlauthorlist}

  \icmlaffiliation{aic}{ETH AI Center, ETH Zurich, Zurich, Switzerland}
  \icmlaffiliation{ibk}{Institute of Structural Engineering, ETH Zurich, Zurich, Switzerland}
  \icmlaffiliation{sdsc}{Swiss Data Science Center, Zurich, Switzerland}

  \icmlcorrespondingauthor{Karin Yu}{karin.yu@ibk.baug.ethz.ch}
  \icmlkeywords{Deep Generative Models, Flow, Formal Grammars, Dynamical Systems, Ordinary Differential Equations, Symbolic Regression, Equation Discovery}

  \vskip 0.3in
]



\printAffiliationsAndNotice{}  % no special notice (required even if empty)

% abstract
\begin{abstract}
  Understanding natural and engineered systems often relies on symbolic formulations, such as differential equations, which provide interpretability and transferability beyond black-box models. 

We introduce Latent Grammar Flow (LGF), a neuro-symbolic generative framework for discovering ordinary differential equations from data. LGF embeds equations as grammar-based representations into a discrete latent space and forces semantically similar equations to be positioned closer together with a behavioural loss. Then, a discrete flow model guides the sampling process to recursively generate candidate equations that best fit the observed data. Domain knowledge and constraints, such as stability, can be either embedded into the rules or used as conditional predictors.
\end{abstract}

% chapters
\section{Introduction}
\label{ch1:intro}

Natural and engineered systems are commonly described by dynamical models, in which governing laws are expressed as differential equations derived from observations. These equations enable predictions by linking estimated model responses to measured data~\cite{zembowicz_discovery_1992,schmidt_distilling_2009}, eventually underpinning the foundation of diverse computational models. The inference of differential equations is a challenging task due to innate noise, uncertainty and ill-posedness. Although successful applications of data-driven methods in simulating dynamical systems exist~\cite{chen_neural_2018}, they are often black-box and neither transfer easily to similar systems nor support clear interpretation, limiting the extraction of further knowledge. Symbolic regression addresses this limitation by discovering compact, transparent mathematical expressions directly from data, providing interpretability, transferability and, hence, an extension to the body of scientific knowledge. \\

\textbf{Symbolic regression} methods can be divided into generative methods, which learn a distribution over equations and provide a symbolic equation given a trajectory, and search methods, which explore the space of expressions using heuristics, or a hybrid thereof. Generative methods sample a complete equation, whose numerical constants are subsequently identified through optimisation. Such learning-based methods rely on a large corpus of data consisting of both the symbolic equations and their numerical realisations and are often based on deep transformer  architectures~\cite{lample_deep_2019,dascoli_deep_2022,kamienny_deep_2023,becker_predicting_2023}. On the other hand, search methods explore the space of potential equations with evolutionary algorithms~\cite{koza_genetic_1992,bongard_automated_2007,tsoulos_solving_2006,cranmer_discovering_2020,cranmer_interpretable_2023}, combinatorial search~\cite{brence_probabilistic_2021,omejc_probabilistic_2024} or reinforcement learning~\cite{petersen_deep_2021,crochepierre_reinforcement_2022}. The search space can thus be restricted through grammars~\cite{hoai_solving_2002,brence_probabilistic_2021,omejc_probabilistic_2024}, embedded directly into the neural network architecture~\cite{sahoo_learning_2018,garmaev_nomto_2025}, or be defined by a candidate dictionary~\cite{brunton_discovering_2016,kaheman_sindy-pi_2020,kacprzyk_d-cipher_2023}. Hybrid models combine learning- and search-based methods by sampling equations with a trained model (e.g. by embedding it into a latent space~\cite{yu_grammar-based_2025,li_gensr_2026} or using a large language model~\cite{novikov_alphaevolve_2025,shojaee_llm-sr_2025}) and adapting the prior (e.g. latent vectors or prompts) via use of a suitable optimisation algorithm often based on evolutionary strategies. 

\textbf{Ordinary differential equations} (ODEs) are typically used to model time- or space-dependent dynamical systems. ODEs can be represented in implicit form, as in \cref{eq1}, or in explicit form, as in \cref{eq2}, where $\mathcal{D}$ denotes the differential operator, $u(t)$ the solution of the system, $t$ the independent variable, $F(t)$ the input (or force) function and $n$ the order of the ODE. 

\begin{align}
    \mathcal{D}(u(t)) & - F(t) = 0, \label{eq1} \\
    u^{(n)}(t) & =f\left(u(t), u'(t),...,u^{(n-1)}(t), F(t)\right) \label{eq2}
\end{align}

Not all implicit ODEs can be globally reformulated in explicit form. Most symbolic regression methods~\cite{becker_predicting_2023,cranmer_interpretable_2023,dascoli_odeformer_2024} approach ODE discovery by casting it as a function approximation task, i.e. learning a mapping $f(\mathbf{X})=\mathbf{y}$ between input data $\mathbf{X}$ and outputs $\mathbf{y}$~\cite{kacprzyk_d-cipher_2023}. Consequently, to avoid trivial equations and enforce the order of the system, these methods typically restrict the formulation to explicit ODEs.

Motivated by the need for interpretable and flexible equation discovery, we propose the \textit{Latent Grammar Flow} (LGF) for the discovery of both explicit and implicit ODEs. LGF embeds the symbolic structure of an equation, represented by a sequence of rules, into a discrete latent space and employs flow-based models together with a decoder to guide sampling within this space. ODEs, like more complex mathematical expressions, are represented symbolically as discrete structures, where transitions between variables and functions (e.g., $\sin$, $\exp$) follow combinatorial rules. Furthermore, expressions must satisfy strict syntactic constraints (e.g. balanced parentheses). Grammars provide a natural way to enforce such structure while offering a concise and compact representation as sequences of rules. 

Existing grammar-based methods, such as ProGED~\cite{omejc_probabilistic_2024,brence_probabilistic_2021} or GODE~\cite{yu_grammar-based_2025}, either rely on Monte-Carlo sampling or a low-dimensional embedding into a continuous latent space. The former can be computationally expensive due to a possibly large search space, while the latter relaxes the discrete nature of the problem, often leading to poorly defined latent representations and increased rejection sampling. In contrast, LGF adopts a discrete dimensionality reduction method and a generative approach with latent flow models, conditioned on domain knowledge. We incorporate knowledge in two ways. First, grammars encode the admissible syntax and structure of mathematical expressions, allowing for nested representations without requiring predefined basis functions (as in dictionary-based methods) or learning syntax from data (as in transformer-based approaches). In this sense, the grammar can be viewed as a structured dimensionality reduction method. Second, flow-based models learn a mapping between a simple and a complex target distribution. This allows us to condition the flow model on domain-specific aspects, such as system order or stability, thereby guiding sampling towards high-probability regions of the search space. \\ \\ \\ 
Our key contributions of LGF can therefore be summarised as:
\begin{itemize}
    \item We propose the Grammar Quantisation Autoencoder (GQAE), embedding a rule-based realisation into a discrete latent space.
    \item We restructure the latent space according to a behavioural distance, capturing not only the structure of equations but also their semantic behaviour.
    \item We incorporate domain knowledge through conditional predictors on the order and stability of ODEs.
    \item We apply a guided sampling strategy in the discrete latent space by coupling discrete flow models with the trained predictors.
    
\end{itemize}
\section{Methodology}
\label{ch3:method}

\subsection{Grammar-based representation}
\label{ch3:method:grammar}

Grammars enable the representation of mathematical expressions as sequences of rules, thereby incorporating structure and syntactic constraints directly within the representation. This eliminates the need to learn token sequences, e.g. through transformers, or to explicitly construct expression trees. A context-free grammar $G$ is represented by the tuple $\{\mathcal{T},\mathcal{N},\mathcal{P},\mathcal{S}\}$~\cite{chomsky_aspects_1969}. $\mathcal{T}$ and $\mathcal{N}$ are distinct sets of terminal and nonterminal symbols, $\mathcal{P}$ is a finite set of production rules and $\mathcal{S}\in\mathcal{N}$ denotes the starting symbol. The map $\upalpha\rightarrow\upbeta$ with $\upalpha\in\mathcal{N}$ and $\upbeta\in(\mathcal{N}\cup\mathcal{T})^*$ forms a production rule $r\in\mathcal{P}$, where * denotes the Kleene star. By recursively applying production rules starting from $\mathcal{S}$, terminal strings $w \in \mathcal{T}^*$ can be generated, forming the language $\mathcal{L}(G) = \{w\in\mathcal{T}^*\mid  \mathcal{S}\rightarrow^*w\}$.  Conversely, a given expression can be mapped to a corresponding sequence of production rules through parsing.

We do not embed scalars as sequences of grammar rules, as this would require several rule applications \cite{yu_grammar-based_2025}, resulting in overly long sequences and more complicated model training. Instead, we introduce the placeholder symbol $C$ to represent scalar constants, whose values are subsequently identified through optimisation (see \cref{ch3:method:optimisation}).

\subsection{Grammar-based discrete latent embedding}
\label{ch3:method:gqae}
\begin{figure*}[h!]
    \centering
    \includegraphics[width=0.9\linewidth]{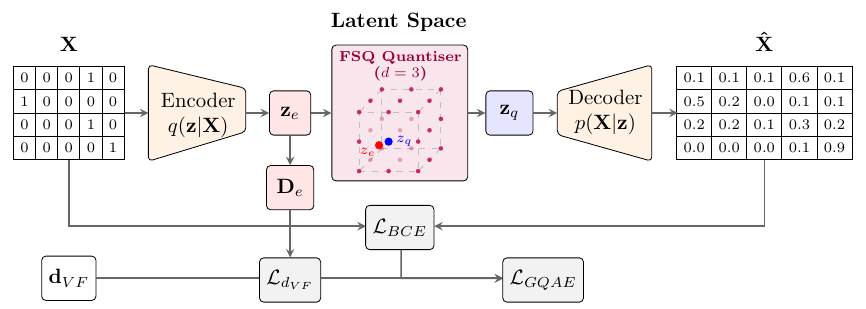}
    \caption{Overview of the training of the GQAE with the semantic loss of the behavioural distance $\mathbf{d}_{VF}$.}
    \label{fig:fig1_methods_overview}
\end{figure*}

We propose the Grammar Quantisation Autoencoder (GQAE). Instead of mapping discrete rules to a continuous latent space, as in the grammar variational autoencoder~\cite{kusner_grammar_2017}, we embed the sequence of rules into a discrete latent space by applying quantisation methods; see \cref{fig:fig1_methods_overview} for an overview. Specifically, we adapt the finite scalar quantisation autoencoder (FSQ-AE) by \citet{mentzer_finite_2023}, which maps the encoded latent vector to a quantised latent vector. For simplicity of training, FSQ-AE is chosen, although other methods such as the vector quantisation variational autoencoder (VQ-VAE)~\cite{oord_neural_2017}, are also applicable. However, VQ-VAE would require a codebook to be trained. Our employed projection into a discrete latent space reduces the dimensionality compared to a continuous latent space, which is in principle unbounded, and further aims to reduce rejection sampling while improving exploration efficiency.

First, the expressions are parsed to extract the sequence of rules, which is represented as a matrix of one-hot encoded vectors $\mathbf{X}\in\{0,1\}^{N_{\max}\times n_r}$ with $N_{\max}$ a user-defined maximum length of the sequence of rules and $n_r$ the number of rules in the grammer $G$. If the sequence length is shorter than $N_{\max}$, the sequence is extended using a dedicated padding rule. The encoder $q(\mathbf{z}\mid \mathbf{X})$ maps this matrix to the latent distribution with size $d \times n_\text{FSQ,cha} \times n_\text{FSQ} $, where $d$ is the dimension of the latent space, $n_\text{FSQ,cha}$ the number of FSQ channels and $n_\text{FSQ}$ the number of FSQ levels. The final equivalent cookbook size is $(n_\text{FSQ})^{n_\text{FSQ,cha}}$. The encoded latent vector $\mathbf{z}_e$ is then quantised with FSQ by assigning the encoded latent vector to the closest quantised vector $\mathbf{z}_q$. Subsequently, the decoder $p(\mathbf{X}\mid \mathbf{z})$ maps the quantised latent vector $\mathbf{z}_q$ to the matrix $\mathbf{\hat{X}}$, from which a sequence of rules is sampled according to the grammar.

\subsection{Behavioural latent distance}
\label{ch3:method:behavioural_distance}

Moving from a continuous variational latent space to a discrete one eliminates the requirement of imposing a Gaussian prior. However, ODEs are not solely defined by their symbolic structure, but also by their induced dynamical behaviour. For example, the solutions of the ODEs $u(t) + 3t,u'(t) = \sin(2t)$ and $u(t) + 3t^2 u'(t) = \sin(2t)$ exhibit markedly different dynamics, despite having similar symbolic structure and corresponding rule sequences. This highlights that structural similarity does not necessarily imply behavioural similarity.

To formalise this notion of behavioural similarity, we build on the work of \citet{meznar_quantifying_2024}, who define a distance between mathematical expressions based on distributions over their behaviour under varying constants, rather than relying on purely structural measures such as the graph edit distance. We adopt a similar perspective and tailor it to the setting of ODEs. To ensure computational tractability, we restrict the evaluation of behaviour to the intrinsic dynamics of the system. Namely, we omit the dependency on the input function $F(t)$ and the initial conditions. Instead, we characterise each equation through the left-hand side of its dynamics (the differential operator) $\mathcal{D}(u(t))$ and evaluate it by sampling the vector field in place of full numerically resolved trajectories. This significantly reduces the computational toll, as it avoids repeatedly solving the ODE for different input functions and initial or boundary conditions.

\cref{tab:vf_sample_points} lists the range and number of samples of the equally spaced input variables (for ODEs up to the second order): $t, u(t)$ and $\dot{u}(t)$. This results in a three-dimensional field of $n_{s,VF} = 8\times16\times16 = 2,048$ data points per sample. 
\begin{table}[]
    \centering
    \caption{Sample points of the vector field.}
    \begin{tabular}{|l|l|l|}
    \hline
        Variable & range & number of samples \\
         \hline
        $t$ & [0,10] & 8 \\
        $u(t)$ & [0,100] & 16 \\
        $\dot{u}(t)$ & [0,500] & 16 \\
        \hline
    \end{tabular}
    \label{tab:vf_sample_points}
\end{table}
For each ODE structure (point in the discrete latent space), the ODE is evaluated 25 times with randomly sampled scalars in the range of $-10$ to $10$ to obtain an approximation of the ODE distribution. The number of samples per ODE, as well as the number of evaluation points, can be increased to obtain a more accurate representation of the induced behaviour, and thus distribution of the ODE. However, this would proportionally increase the computational cost. Based on these sampled evaluations, the mean and standard deviation are computed at each sample point. To ensure numerical stability, the mean is clipped to $\pm10^{10}$ and the standard deviation at $10^{10}$. 

To quantify the distances between the distributions of the ODE structures, we take the logarithm of the mean of the element-wise squared sum of the differences of the means and of the standard deviations of the two distributions according to \cref{{eq:behavioural}}. The logarithm reduces the influence of outliers. 

\begin{align}
d_{VF}& (\boldsymbol{\upmu}_1,\boldsymbol{\upsigma}_1,\boldsymbol{\upmu}_2,\boldsymbol{\upsigma}_2) = \max\Biggl(\log\Bigl(\frac{1}{n_{s,VF}} \nonumber \\
    & \cdot\sum_{i=1}^{n_{s,VF}} \bigl((\upmu_{1,i}-\upmu_{2,i})^2+ (\upsigma_{1,i}-\upsigma_{2,i})^2\bigl)\Bigl),0\Biggl)\label{eq:behavioural}
\end{align}

The behavioural loss can then be formulated as follows:

\begin{align}
\mathcal{L}_{d_{ij}}& (\mathbf{X}_i, \mathbf{X}_j, \mathbf{z}_{e,i}, \mathbf{z}_{e,j})  = \bigl(||\mathbf{z}_{e,i}-\mathbf{z}_{e,j}||_2 \nonumber \\
& -\uplambda_{VF} d_{VF}(\boldsymbol{\upmu}(\mathbf{X}_i),\boldsymbol{\upsigma}(\mathbf{X}_i), \boldsymbol{\upmu}(\mathbf{X}_j),\boldsymbol{\upsigma}(\mathbf{X}_j))\bigl)^2\\
\mathcal{L}_{d_{VF}} & = \frac{1}{n_B\cdot n_B}\sum_{i=1}^{n_B}\sum_{j=1}^{n_B} \mathcal{L}_{d_{ij}}(\mathbf{X}_i, \mathbf{X}_j, \mathbf{z}_{e,i}, \mathbf{z}_{e,j})\label{eq:behavioural_loss}
\end{align}

where $\mathbf{X}_i$ is the matrix of one-hot encoded vectors of the $i$th sample, $\mathbf{z}_{e,i}$ the corresponding encoded latent vector and $\uplambda_{VF}$ a scaling weight equal to $10^{-2}$ and $n_B$ the batch size. $\boldsymbol{\upmu}(\mathbf{X}_i)$ and $\boldsymbol{\upsigma}(\mathbf{X}_i)$ represent the mean and standard deviation of the sampled vector fields of the equation generated by the sequence of rules embedded in $\mathbf{X}_i$. The behavioural loss aims to place semantically similar equations closer together. Both the logarithm and the scaling weight make the restructuring of the latent space less dependent on the specific magnitude of the behavioural distance.

Hence, the GQAE can be trained with the following loss:
\begin{equation}
    \mathcal{L}_{GQAE} = \mathcal{L}_{BCE} + \upbeta_{VF} \mathcal{L}_{VF} \label{eq:gqae_loss}
\end{equation}
where $\mathcal{L}_{BCE}$ is the binary cross-entropy loss with a Sigmoid layer between the ground truth matrix $\mathbf{X}$ and the predicted $\mathbf{\hat{X}}$ of the one-hot encoded vectors, representing the reconstruction loss, and $\upbeta_{VF}=10^{-4}$ is a weight factor to balance the objectives. The Adam optimiser \cite{kingma_adam_2014} is used with the initial learning rate $10^{-3}$ and a metrics-based learning rate scheduler using factor $0.9$, patience $500$ and minimum learning rate of $10^{-4}$ in PyTorch \cite{ansel_pytorch_2024}.

\subsection{Discrete flow matching and conditional guidance}
\label{ch3:method:discrete_flow}

The effective discovery of ODEs does not only aim to infer an accurate equation, but also to incorporate certain domain information and thus constraints. Here, we apply discrete guidance, developed by \citet{nisonoff_unlocking_2025} for both diffusion and flow-based models, to the latent space of GQAE and extend discrete guidance to multiple predictors. We use the discrete flow model \cite{campbell_generative_2024} to sample from the latent space and inject guidance into the discrete latent distribution $p(\mathbf{z}_q)$. We shortly summarise the method of discrete guidance for discrete flow models and its extension, but refer to the original publications for details.

Flow models transform a simple probability distribution $p_0(x_0)$ (e.g. noise) into a complex target distribution $p_1(x_1)$, endowing them with generative capability. In flow matching, this path is specified via a conditional probability flow $p_{t\mid 1}(x_t\mid x_1)$, whose marginal distribution is obtained by integrating over the data distribution $p_1(x_1)$. In continuous state-space models, a score function is learned, which steers the denoising process. For discrete state spaces, however, gradient-based score formulations are not directly applicable. Instead, discrete flow matching can be realised with continuous time Markov chains, where a rate matrix $R_t$ is related to the change in $p_t$ according to the Kolmogorov equation and encodes whether the state stays unchanged or jumps to a different state \cite{campbell_generative_2024}. A neural network $p_{1\mid t}^\uptheta(x_1\mid x_t)$, termed the denoising model in \citet{nisonoff_unlocking_2025}, is trained to map noisy states $x_t$ to target states $x_1$, from which a corresponding rate matrix $R_t$ can be derived. Thus, the discrete flow model learns a conditional flow, defining a time-dependent rate matrix. This matrix steers the sampling towards the target distribution to increase the probability of transitioning to the final target state. \citet{nisonoff_unlocking_2025} propose to integrate the guidance into the conditional rate matrix through the predictor model $p(y\mid x,t)$:
\begin{equation}
    R_t(x,\tilde{x}\mid y) = \frac{p(y\mid \tilde{x},t)}{p(y\mid x,t)}R_t(x,\tilde{x}) \label{eq:guidance}
\end{equation}
which can be approximated as follows
\begin{align}
    \log\left(\frac{p^\upphi(y\mid \tilde{x},t)}{p^\upphi(y\mid x,t)}\right) & = \log\left(p^\upphi(y\mid \tilde{x},t)\right) \nonumber \\
    & \quad - \log\left(p^\upphi(y\mid x,t)\right),
\end{align}
where $p^\upphi(y\mid x,t)$ is the learned predictor model. As the guidance by \citet{nisonoff_unlocking_2025} is only defined for one guidance predictor, we extend this to multiple predictors with the joint probability: 
\begin{equation}
    p^{\boldsymbol{\upphi}}(\mathbf{y}\mid x,t) \propto \prod_{i=1} p^{\upphi_i}(y_i,x,t)
\end{equation}

where $n_p$ is the total number of predictors. The predictor variables $\{y_i\}_{i=1}^{n_p}$ are assumed to be conditionally independent given $x$ and $t$. Hence, the discrete flow model is trained from $\mathbf{z}_q$ to $\mathbf{z}_d$, which is then accepted by the decoder $p(\mathbf{X}\mid \mathbf{z})$. 

\subsection{Optimisation, discovery process and guidance}
\label{ch3:method:optimisation}

LGF employs a two-stage nested discovery process, where, in the outer stage, GQAE samples an ODE skeleton and, in the inner stage, the Nelder-Mead algorithm~\cite{nelder_simplex_1965} implemented in Scipy~\cite{virtanen_scipy_2020} optimises the scalar values of the expression. We first describe the outer optimisation scheme with GQAE and then dive into the details of the inner scalar optimisation process.

Once the GQAE, the denoiser, and the offline (static) predictors have been trained, the skeleton expression can be discovered. Static predictors encode properties that remain invariant across problem instances, such as the order of the ODE. In contrast, dynamic predictors capture problem-dependent characteristics, including stability and the optimisation objective value (after scalar identification), which must be evaluated anew for each dataset.

In particular, stability plays a central role in the inference process. Many physical and engineered systems exhibit stable behaviour in their observed responses, reflecting underlying dissipative or bounded dynamics. Consequently, incorporating stability as a guiding criterion aligns the discovery process with domain knowledge, promoting the identification of physically meaningful and plausible equations.

To assess the stability of candidate ODEs, we first apply Lyapunov’s indirect method via linearisation~\cite{lyapunov_general_1992}. In particular, each ODE is converted into its equivalent first-order state-space form, and local stability is inferred from the eigenvalues of the Jacobian of the resulting vector field evaluated at equilibrium points.
In our formulation, stability is assessed based solely on the intrinsic dynamics defined by the differential operator $\mathcal{D}(u(t))$, i.e. the left-hand side of the ODE, as the input function $F(t)$ and initial or boundary conditions are not considered. This operator is interpreted as inducing the underlying vector field of the system. However, Lyapunov’s indirect method is generally inconclusive for non-autonomous systems, where the vector field explicitly depends on time $t$~\cite{khalil_nonlinear_2002}.
In such cases, we approximate stability numerically by analysing the boundedness and decay properties of the induced dynamics. Specifically, we evaluate the behaviour of the operator over sampled states and assess whether the resulting vector field exhibits dissipative tendencies. While this does not constitute a formal stability proof, it provides a practical criterion consistent with the reduced representation employed in our framework.

Symbolic regression methods do not solely aim to extract accurate expressions but also to ensure interpretability, which is often associated with lower model complexity. Accordingly, we define the optimisation objective as a weighted sum of the accuracy $\mathcal{L}_\text{accuracy}$ and complexity $\mathcal{C}$ of the expression (defined as the total number of operations, variables and constants), where $n_s$ is the number of samples and $\upalpha$ the weight, typically chosen as $\upalpha=0.1$~\cite{yu_grammar-based_2025}:

\begin{equation}
    \mathcal{L}_{IC} = \upalpha\mathcal{C}+(1-\upalpha)n_s\mathcal{L}_\text{accuracy}
\end{equation}

The outer optimisation loop follows a sequential sampling and refinement process, where an initial population $n_\text{pop,0}$ is first sampled. For this population, the dynamic conditions are evaluated and the dynamic predictors are trained. Subsequently, iterative stages of sampling, evaluating a new population $n_{pop}$, and training of dynamic predictors on the entire set of expressions and evaluations sampled so far follow. The procedure terminates either when the maximum number of iterations $i_{\max}^{\text{out}}$ is reached or when the convergence criteria are satisfied. Convergence is defined in terms of both exploratory capability and improvement of the optimisation objective. Specifically, termination occurs when the number of newly discovered unique expressions falls below a threshold, i.e.
\begin{equation}
(n_{\text{unique},i} - n_{\text{unique},i-1}) \leq \varepsilon_{\text{unique}} \, n_{\text{pop}},
\end{equation}
and the optimisation objective stagnates, measured by the mean top-\textit{k} objective values satisfying
\begin{equation}
\mathcal{L}_{IC,k,i} \geq (1 - \varepsilon_{IC,k}) \, \mathcal{L}_{IC,k,i-1}.
\end{equation}

As the optimisation landscape is non-smooth and gradients of the solution are not readily available, the derivative-free Nelder-Mead algorithm is applied to the inner scalar optimisation, performed in two stages. In the first stage, the tolerances $\texttt{fatol}=0.1$ and $\texttt{xatol}=0.5$ are employed, with a maximum of $i_\text{max}^\text{in1}=50$ iterations. If the resulting loss falls below a threshold of $25$, a second refinement stage is initiated with tighter tolerances $\texttt{fatol}=0.001$ and $\texttt{xatol}=0.1$ for $i_\text{max}^\text{in2}$ iterations. This two-stage scalar optimisation speeds up the process and reduces the time spent on less promising skeleton expressions compared to a single-stage one with the same maximum number of iterations $\left(i_\text{max}^\text{in1}+i_\text{max}^\text{in2}\right)$. The equations are evaluated either through the ODE residual $\mathcal{L}_{DE}$, which is the mean squared error of the ODE evaluation, or the solution loss $\mathcal{L}_{SOL}$, which computes the standardised mean squared error of all numerical trajectories and requires solving the ODE. $\mathcal{L}_\text{accuracy}$ is either defined as $\mathcal{L}_{DE}$ or, if $\mathcal{L}_{DE}$ is below the threshold $\uptheta_{DE}=200$, as $\mathcal{L}_{SOL}$. This strategy avoids the computational cost of solving ODEs for poorly performing candidate expressions.

To promote further simplifications of the expressions, summation terms or in some cases also multiplication terms are eliminated, if their squared sum is below $\upvarepsilon_\text{sum}=10^{-2}$, respectively, summed absolute error from 1 below $\upvarepsilon_\text{mul}$ (here $\upvarepsilon_\text{mul} = \upvarepsilon_\text{sum}$ is assumed). 

\section{Results}
\label{ch4:results}

We demonstrate the efficacy and increased sample efficiency of LGF on three benchmarks. The benchmarks are compared according to the relative L2 error, which is defined as follows:
\begin{equation}
    \text{Relative L2 error}(\mathbf{y},\hat{\mathbf{y}}) = \frac{||\mathbf{y}-\hat{\mathbf{y}}||_2}{||\mathbf{y}||_2}
\end{equation}
For the averaged evaluations, relative L2 errors above 1 are floored to prevent outliers from skewing the statistics too significantly. For all benchmarks, the assumed observed trajectory is corrupted with 5\% Gaussian noise. As most methods, aside from ODEFormer, require all modes, the remaining modes are approximated with a multilayer perceptron. Details of the benchmarks and their results can be found in \cref{app:benchmarks}.

\begin{figure}
    \centering
    \includegraphics[scale = 0.6]{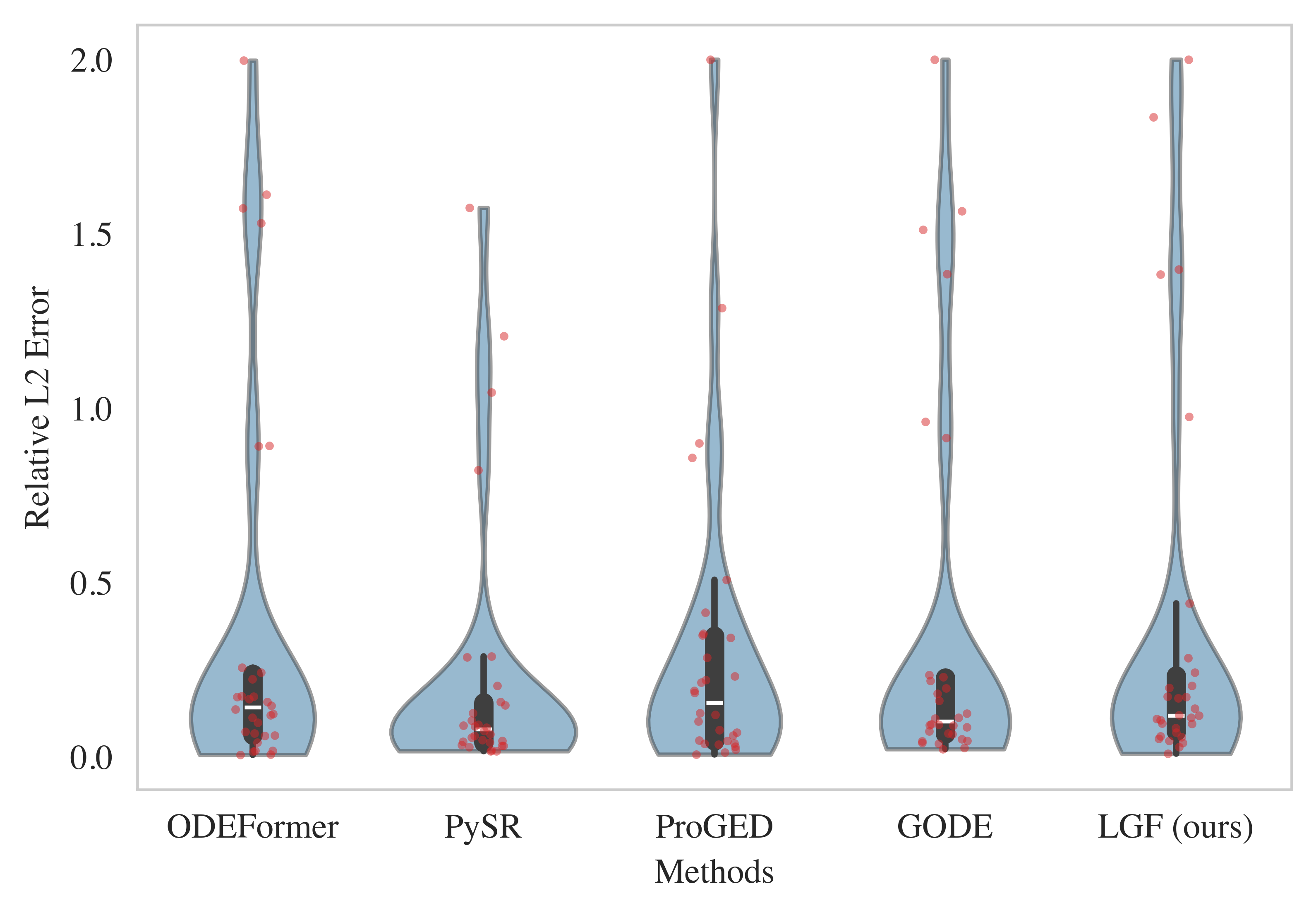}
    \caption{Violin plots of Benchmark 1, where the distribution is shown in blue and single values as red dots.}
    \label{fig:b1_results_violin}
\end{figure}

Benchmark 1 contains 30 explicit one-dimensional first-order ODEs, of which 23 are from the ODEBench by \citet{dascoli_odeformer_2024} and seven from \citet{yu_grammar-based_2025}. The noisy observed data is $\tilde{u}_t$. LGF assumes $i_\text{max}^\text{out} = 10$ iterations of the outer optimisation loop (including the initialisation) with a population size of $n_\text{pop}=100$. As this benchmark only contains explicit first-order ODEs, neither the order nor the stability predictors (i.e. guidance) was applied. Hence, the generative process was guided by the optimisation objective. LGF is compared with the transformer-based generative method ODEFormer \cite{dascoli_odeformer_2024}, the evolutionary algorithm-based search method PySR \cite{cranmer_interpretable_2023}, the grammar-based search method ProGED \cite{brence_probabilistic_2021,omejc_probabilistic_2024} and the hybrid grammar-based method GODE \cite{yu_grammar-based_2025}. A description of these methods can be found in \cref{app:benchmarks}. The resulting mean relative L2 errors are listed in \cref{tab:b1_results}, including the mean and standard deviation (std) of the complexities. \cref{fig:b1_results_violin} depicts the distribution of the relative L2 errors in violin plots. Benchmark 1 shows that PySR is the best model in terms of accuracy closely followed by ProGED; however, LGF performs similarly to ODEFormer and GODE. Moreover, the mean complexity of ODEFormer is significantly lower than those of the other methods, indicating that ODEFormer favours simpler expressions and might not capture the true dynamics well. In contrast, the mean complexities of ProGED, GODE and LGF are fairly close to that of the ground truth, whereas the one of PySR differs by one standard deviation.

\begin{table}[h!]
    \centering
    \caption{Mean relative L2 errors (bold best and underlined second best) and complexity for different models of Benchmark 1.}
    \begin{tabular}{|l|r|r|r|}
    \hline
        Mean relative & $u_t$ & $\dot{u}_t$ & Complexity \\ L2 error  & & & (mean$\pm$std)\\
        \hline
       Ground truth & - & - & $23.2\pm8.6$ \\
       ODEFormer & 0.148 & 0.224 & $11.9\pm5.8$\\
       PySR & \textbf{0.084} & \textbf{0.148} & $17.9\pm5.5$ \\
       ProGED & \underline{0.095} & 0.211 & $25.6\pm0.9$\\
       GODE & 0.150 & 0.211 & $25.6\pm9.9$\\
       LGF & 0.139 & \underline{0.221} & $27.2\pm9.8$ \\
       \hline
    \end{tabular}
    \label{tab:b1_results}
\end{table}

Benchmark 2 from \citet{yu_grammar-based_2025} contains five linear and five nonlinear ODEs. The noisy observed data is $\tilde{u}_t$ and the order and stability properties are assumed to be known. Ten iterations of the outer optimisation loop were assumed for LGF with a population size of $n_\text{pop}=100$. The mean relative L2 errors and mean and standard deviation of the complexity metric are shown in \cref{tab:b2_results}. LGF is compared with PySR, ProGED and GODE. It beats all other methods in terms of the mean relative L2 errors of $u_t$ and $\dot{u}_t$. LGF places second, after GODE in terms of $\ddot{u}_t$, while using a population size half that of GODE, underlining its increased sampling efficiency. Contrary to Benchmark 1, the mean complexity of PySR is the highest, followed by LGF, although LGF shows higher scatter.

\begin{table}[h!]
    \centering
    \caption{Mean relative L2 errors and complexity for different models of Benchmark 2.}
    \begin{tabular}{|l|r|r|r|r|}
    \hline
        Mean relative & $u_t$ & $\dot{u}_t$ & $\ddot{u}_t$ & Complexity \\ L2 error & & & & (mean$\pm$std)\\
        \hline
        Ground truth & - & - & - & $18.1\pm4.8$ \\
       PySR & \underline{0.115} & 0.145 & 0.379 & $29.5\pm7.0$ \\
       ProGED & 0.326 & 0.406 & 0.811 & $17.2\pm3.5$ \\
       GODE & 0.120 & \underline{0.133} & \textbf{0.187} & $18.1\pm4.9$\\
       LGF & \textbf{0.110} & \textbf{0.112} & \underline{0.309} & $23.2\pm11.8$ \\
       \hline
    \end{tabular}
    \label{tab:b2_results}
\end{table}

Benchmark 3 contains seven dynamical systems, ranging from the linearised pendulum, over the Van der Pol oscillator to an exponential stiffness. This benchmark shows a partial observable system with noise, where the noisy measurements are the accelerations $\tilde{\ddot{u}}_t$, as this mode is often measured in real-life structural dynamics problems. All examples are one-dimensional, of second order and considered stable according to the stability assessment in \cref{ch3:method:optimisation}. Benchmark 3 shows the comparison between PySR and LGF, as PySR is a widely popular function approximator that has shown favourable performances in the previous two benchmarks.  The order, stability property and input function are assumed to be given, representing a case of partially available domain knowledge. The order can be induced into the PySR framework by formulating the function approximation task accordingly. However, there is no simple way to check stability and induce the input function, while avoiding trivial equations. LGF assumes $i_\text{max}^\text{out}=10$ iterations of the outer optimisation loop with a population size of $n_\text{pop}=250$.

\begin{table}[h!]
    \centering
    \caption{Relative L2 errors and complexities of the predicted ODEs by PySR and LGF for all ODE examples of Benchmark 3.}
    \begin{tabular}{|l|r|r|r|r|}
    \hline
        Mean relative & $u_t$ & $\dot{u}_t$ & $\ddot{u}_t$ & Complex. \\ 
        L2 error & & & & \\
        \hline
        \textit{Pendulum} & & & & 25 \\
       PySR & 3.707 & 0.389 & 0.146 & 25 \\
       LGF & \textbf{0.017} & \textbf{0.008} & \textbf{0.003} & 27 \\
       \hline
        \textit{Duffing oscil.} & & & & 31 \\
       PySR & 1.455 & 1.492 & 1.600 & 32 \\
       LGF &\textbf{ 0.278} & \textbf{0.237} & \textbf{0.288} & 35 \\
       \hline
        \textit{Van d. Pol oscil.} & & & & 29 \\
       PySR & 20.517 & 3.904 & 0.989 & 38 \\
       LGF & \textbf{0.338} & \textbf{0.275} & \textbf{0.406} & 31 \\
       \hline
        \textit{Nonlin. damp. 1} & & & & 37 \\
       PySR & 0.557 & 0.575 & 0.551 & 51 \\
       LGF & \textbf{0.325} & \textbf{0.334} & \textbf{0.484} & 42 \\
       \hline
        \textit{Nonlin. damp. 2} & & & & 37 \\
       PySR & 6.410 & 2.092 & 1.261 & 41 \\
       LGF & \textbf{0.171} & \textbf{0.273} & \textbf{0.549} & 35 \\
       \hline
        \textit{Damped oscil.} & & & & 36 \\
       PySR & \textbf{0.102} & \textbf{0.154} & \textbf{0.162} & 28 \\
       LGF & 0.143 & 0.168 & 0.225 & 42 \\
       \hline
        \textit{Exp. stiffness} & & & & 29 \\
       PySR & 0.405 & 0.526 & 0.504 & 37 \\
       LGF & \textbf{0.009} & \textbf{0.010} & \textbf{0.012} & 31 \\
       \hline
    \end{tabular}
    \label{tab:b3_results}
\end{table}

In most ODE examples, our proposed method LGF outperforms PySR in terms of accuracy. The main reason is that PySR operates directly on the explicit ODE form as a function approximator. Consequently, PySR neither easily allows for imposing the input function nor solves the discovered ODEs. Skipping the ODE solving step makes PySR significantly faster. However, its predictions can be misleading: PySR may provide seemingly low losses without verifying whether the solutions of the proposed ODEs match the ground truth. The residual loss of an ODE can be low while its solution is still far from the ground truth. By incorporating the input function and solving the ODEs, LGF can discover more accurate and meaningful ODEs.

\section{Conclusion}
\label{ch5:conclusion}

The discovery of symbolic ODEs poses a complex task due to noise, uncertainty and inherent ill-posedness. In this work, we introduce Latent Grammar Flow (LGF), a hybrid neuro-symbolic framework that combines structured representations with generative modelling for the inference of implicit and explicit one-dimensional ODEs. First, we propose the Grammar Quantisation Autoencoder (GQAE), which embeds symbolic expressions into a discrete latent space. Second, we apply the discrete flow model to guide sampling within this discrete latent space. This enables the incorporation of domain knowledge, such as system order and stability, directly into the discovery process, aligning the discovered equations with physically meaningful behaviour. A key advantage of LGF lies in its ability to integrate expert knowledge while remaining data-driven, and in its computational efficiency compared to transformer- or large language model-based approaches, which often require substantially greater resources. Although LGF is here validated on first- and second-order single degree of freedom ODEs, it could be easily extended to higher dimensional systems by adjusting the respective optimisation objective function (accuracy loss function). In addition, LGF can be generalised to systems of ODEs through appropriate grammar construction. Future work will explore the incorporation of additional dynamical properties, such as autonomy, as well as broader classes of systems. 

\bibliography{references}
\bibliographystyle{icml2026}

%%%%%%%%%%%%%%%%%%%%%%%%%%%%%%%%%%%%%%%%%%%%%%%%%%%%%%%%%%%%%%%%%%%%%%%%%%%%%%%
%%%%%%%%%%%%%%%%%%%%%%%%%%%%%%%%%%%%%%%%%%%%%%%%%%%%%%%%%%%%%%%%%%%%%%%%%%%%%%%
% APPENDIX
%%%%%%%%%%%%%%%%%%%%%%%%%%%%%%%%%%%%%%%%%%%%%%%%%%%%%%%%%%%%%%%%%%%%%%%%%%%%%%%
%%%%%%%%%%%%%%%%%%%%%%%%%%%%%%%%%%%%%%%%%%%%%%%%%%%%%%%%%%%%%%%%%%%%%%%%%%%%%%%
\newpage
\appendix
\onecolumn

\section*{Appendix to: Neuro-Symbolic ODE Discovery with Latent Grammar Flow}
\section{Details on the benchmarks}
\label{app:benchmarks}

This appendix presents the implementation details of the benchmarks. Common settings across all three benchmarks are listed first. To maintain a fair comparison among the methods, the same grammars and datasets as in \citet{yu_grammar-based_2025} are used for each benchmark according to the GitHub repository.

Certain hyperparameters of the training of the GQAE model as well as the flow model were kept consistent across all three benchmarks: three layers of convolutional neural network with kernel sizes 7, 8 and 9 and ReLU activation function, followed by two residual layers of dimension $n_{res}=60$ for the encoder, $n_\text{FSQ,cha}=4$ layers of $n_\text{FSQ}=9$ FSQ levels for the quantisation method, and, a hidden layer of dimension $80$ with the activation function Exponential Linear Unit (ELU) \cite{clevert_fast_2015} and two bidirectional GRU layers with dimension of $80$ for the decoder. The scaling weight of the Wasserstein distances was $\uplambda_{VF} = 10^{-2}$. The batch size $n_B$ was $512$ and early stopping was applied after $2,000$ epochs. The main hyperparameters of GQAE, which required tuning for each new dataset and grammar, were the latent dimension $d$ and the number of FSQ levels $n_{FSQ}$. 

For the discrete flow model, the denoising model had two hidden layers of dimension 5,000 with ReLU activation and $20\%$ dropout. The denoising model was trained for 300 epochs with the Adam optimiser \cite{kingma_adam_2014} and learning rate $10^{-3}$. The same hyperparameters were selected for all predictors: one hidden layer with 500 neurons with ELU activation and $20\%$ dropout, trained for 200 epochs with the Adam optimiser and learning rate $10^{-3}$. The order and stability conditions were treated as categorical variables, and the optimisation objective as a continuous one. Additionally, the length of the sequence $n_\text{shape}$ and parameter $S$ need to be defined for each model, depending on the latent dimension and number of FSQ levels. For all three benchmarks, parameter $S$ is $5$. For the generation with the discrete flow model, the hyperparameters were noise $\upeta=10$, temperature of $x_1=1$, time step of $0.01$ and guide temperature of $1$. As the sampling process requires numerous function calls to the predictor model, the Taylor-approximated guidance is applied. For the convergence criteria, top-\textit{k} was equal to $10\%$ of the population size $n_\text{pop}$ and the tolerances were $\upvarepsilon_{IC,k}=0.01$ and $\upvarepsilon_{unique}=0.1$.

The benchmark-dependent hyperparameters are either listed in \cref{tab:app_hyperparam} or described in the respective subsections.
\begin{table}[h!]
    \centering
    \caption{Overview of the varying hyperparameters of LGF across the three benchmarks.}
    \begin{tabular}{|l|l|l|l|}
    \hline
    Hyperparameter & Benchmark 1 & Benchmark 2 & Benchmark 3 \\
    \hline
        $d$ (GQAE) & 17 & 24 & 17\\
        $i_\text{max}^\text{in2}$ (Nelder-Mead) & 150 & 250 & 250 \\
        $n_\text{shape}$ (discrete flow model) & 68 & 96 & 68\\
        \hline
    \end{tabular}
    \label{tab:app_hyperparam}
\end{table}

\subsection{Benchmark 1}

To ensure consistent comparison with existing symbolic regression methods, the evaluations from \citet{yu_grammar-based_2025} were taken. A short description of all existing symbolic regression methods follows; for further implementation details, we refer to the reference.

ODEFormer \cite{dascoli_deep_2022} is a deep transformer-based generative symbolic regression method. It learns an operator between the numerical trajectory and the mathematical expression and therefore requires substantial computing resources to generate the training dataset ($\sim$50 million samples) and train the large-scale model (with 60.7 million parameters). Although ODEFormer can be applied as a purely generative symbolic regression model, in practical discovery tasks, beam search (with a beam size of $50$ and a temperature of $0.1$) is used to find better fitting equations.

PySR \cite{cranmer_interpretable_2023} is a fast, high-performance symbolic regression method exploiting a multi-population evolutionary algorithm. Technically, for this benchmark, PySR could explore $20\cdot20\cdot30\cdot380=4,560,000$ skeleton expressions without specifying the number of scalar optimisation iterations. However, it terminated earlier in all cases. 

\begin{figure}[h!]
    \centering
    \includegraphics[scale = 0.68]{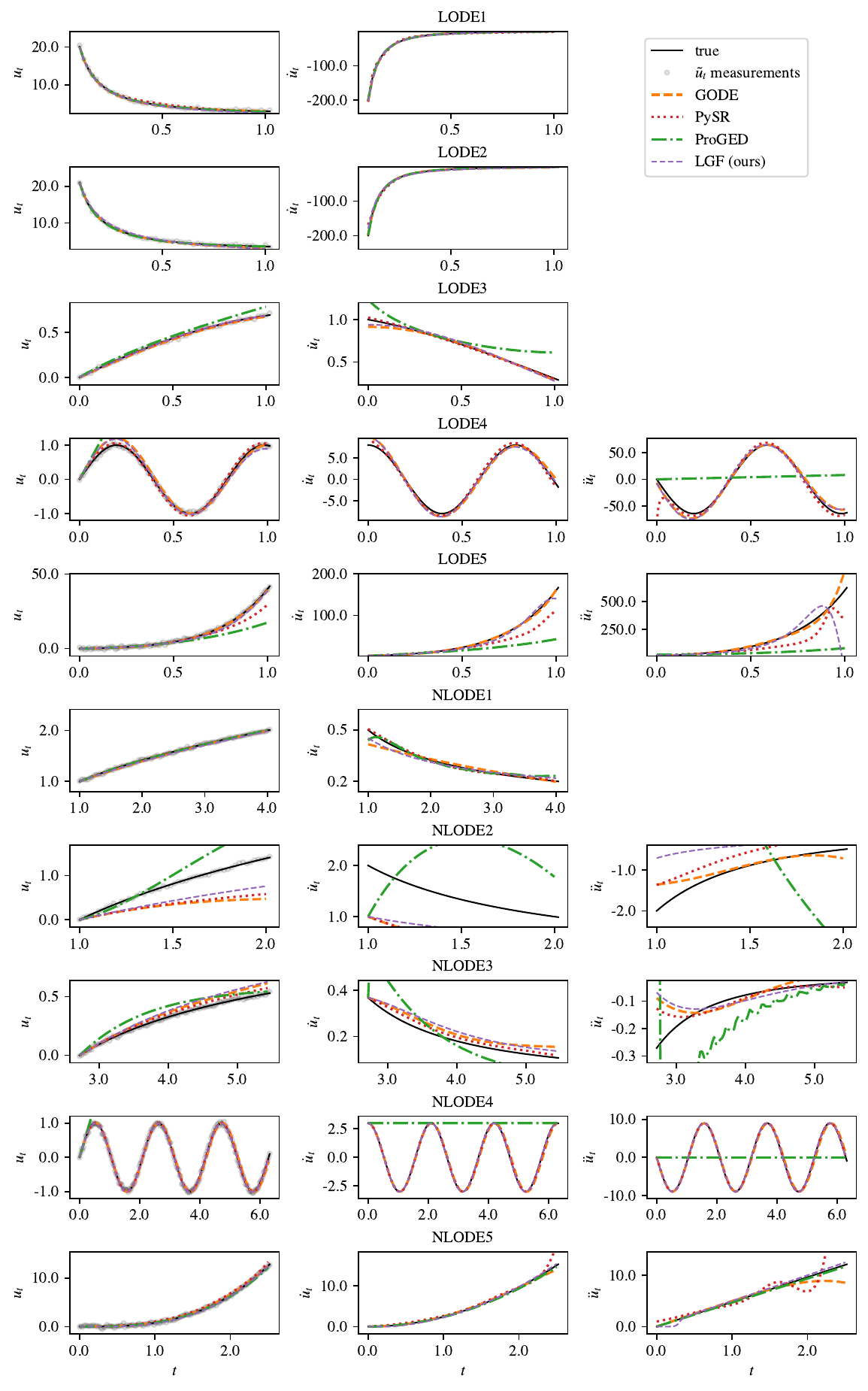}
    \caption{The numerical trajectories of the ground truth and predicted ODEs of Benchmark 2.}
    \label{fig:app_B2_num}
\end{figure}

ProGED \cite{brence_probabilistic_2021,omejc_probabilistic_2024} is a probabilistic grammar-based search method using Monte Carlo sampling. ProGED could technically explore $100\cdot100=10,000$ skeletons without a specified number of scalar optimisation runs in Benchmark 1. 

GODE \cite{yu_grammar-based_2025} is a hybrid grammar-based method, which similarly as our method represents ODEs as sequences of rules but uses the grammar variational autoencoder~\cite{kusner_grammar_2017} to embed the sequences of rules into a continuous latent space. The continuous latent space is then explored with a gradient-free evolution strategy. For Benchmark 1, GODE could explore up to $10\cdot100=1,000$ skeletons with a maximum of $300$ function evaluations per skeleton expression.

For LGF, the additional hyperparameters are listed in \cref{tab:app_hyperparam}. In this benchmark, LGF was set to explore up to $i_\text{max}^\text{out}\times n_\text{pop}=10\times100=1,000$ skeletons with up to $200$ scalar optimisation iterations. For the simplification of the equations, only the summation terms were considered. 

\subsection{Benchmark 2}

To ensure consistent comparison with existing symbolic regression methods, the evaluations from \citet{yu_grammar-based_2025} were taken. For ProGED and PySR, the same settings as in Benchmark 1 are selected. While GODE searched at maximum $10\cdot200=2,000$ skeletons with $500$ function evaluations per scalar optimisation.

For LGF, the additional hyperparameters are listed in \cref{tab:app_hyperparam}. LGF could explore up to $i_\text{max}^\text{out}\times n_\text{pop}=10\times100=1,000$ skeletons with up to $300$ scalar optimisation iterations. For the simplification of the equations, both summation and multiplication terms were considered. 

\cref{fig:app_B2_num} shows the displacement $u_t$, velocity $\dot{u}_t$ and acceleration $\ddot{u}_t$, if applicable, for the ground truth and predicted ODEs of Benchmark 2 for the results of PySR, GODE, ProGED and LGF.

\subsection{Benchmark 3}
Benchmark 3 includes four additional dynamical systems compared to the third benchmark in \citet{yu_grammar-based_2025}. All equations including the domain $t$, sampling frequency $f_s$ and initial conditions $[u(t_0),\dot{u}(t_0)]$ are listed in \cref{tab:app_B3_odes}.

\begin{table}[h!]
    \centering
    \renewcommand{\arraystretch}{1.5}
    \caption{Details of ODEs of Benchmark 3.}
    \begin{tabular}{|l|l|l|l|l|}
    \hline
        Problem & ODE & $t$ & $f_s$ & $[u(t_0),\dot{u}(t_0)]$ \\
        \hline 
        Pendulum & $2\frac{d^2}{dt^2}u(t)+\frac{d}{dt}u(t)+5u(t)=2\sin(0.5t)$ & $[0,60]$ & $10$ & $[0.0,3.0]$ \\
        Duffing oscil. & $5\frac{d^2}{dt^2}u(t)+\frac{d}{dt}u(t)+7u(t)+25u^3(t)=\cos(2t)$ & $[0,30]$ & $10$ & $[0.0,1.5]$\\
        Van d. Pol oscil. & $\frac{d^2}{dt^2}u(t)+5\frac{d}{dt}u(t)\left(1-u^2(t)\right)+u(t)=\cos(2t)$ & $[0,30]$ & $10$ & $[1.0,0.0]$ \\
        Nonlin. damp. 1 & $3\frac{d^2}{dt^2}u(t)+1.3\frac{d}{dt}u(t)+2.5\left(\frac{d}{dt}u(t)\right)^2+4u(t)=1.5\cos(\frac{2}{3}t)$ & $[0,30]$ & $15$ & $[1.0,0.2]$ \\
        Nonlin. damp. 2 & $2\frac{d^2}{dt^2}u(t)+2.7\frac{d}{dt}u(t)+3\left(\frac{d}{dt}u(t)\right)^3+2u(t)=3\sin(0.7t)$ & $[0,30]$ & $15$ & $[2.0,0.0]$ \\
        Damped oscil. & $1.2\frac{d^2}{dt^2}u(t)+2.2\frac{d}{dt}u(t)+0.8t\frac{d}{dt}u(t)+3u(t)=2.3\sin(1.2t)$ & $[0,30]$ & $20$ & $[0.4,0.2]$ \\
        Exp. stiff. & $\frac{d^2}{dt^2}u(t)+1.2\frac{d}{dt}u(t)+3e^{0.25t}u(t)=1.8\cos(1.5t)$ & $[0,20]$ & $20$ & $[0.5,0.8]$ \\
        \hline
    \end{tabular}
    \label{tab:app_B3_odes}
\end{table}

As the benchmark has been extended, the hyperparameters of PySR have been modified as listed in \cref{tab:app_pysr_b3}, which differ from the default settings. Technically, if PySR did not terminate the search early, it could explore $60\cdot20\cdot30\cdot500=18,000,000$ skeleton expressions, not accounting for the scalar optimisation part, which would result in an even higher number of function evaluations. Yet, it typically terminated earlier with a substantially lower but unknown number of function evaluations.

\begin{table}[h!]
    \centering
    \caption{Hyperparameters of PySR for Benchmark 3.}
    \begin{tabular}{|l|l|}
    \hline
        Setting &  \\
        \hline
        \texttt{niterations} & 60 \\ 
        \texttt{populations} & 20 \\ 
        \texttt{population\_size} & 30 \\ 
        \texttt{ncycles\_per\_iteration} & 500 \\ 
        \texttt{binary\_operators} & [+,*,/,-,\^{}] \\ 
        \texttt{unary\_operators} & [$\cos,\exp,\sin,\log$] \\ 
        \texttt{model\_selection} & best \\ 
        \texttt{elementwise\_loss} & \verb|loss(prediction, target) = (target-prediction)^2| \\ 
        \hline
    \end{tabular}
    \label{tab:app_pysr_b3}
\end{table}

For LGF, the additional hyperparameters are listed in \cref{tab:app_hyperparam}. LGF could explore up to $i_\text{max}^\text{out}\times n_\text{pop}=10\times250=2,500$ skeletons with up to $300$ scalar optimisation iterations. For the simplification of the equations, both summation and multiplication terms were considered. 

\begin{figure}[h!]
    \centering
    \includegraphics[scale = 0.73]{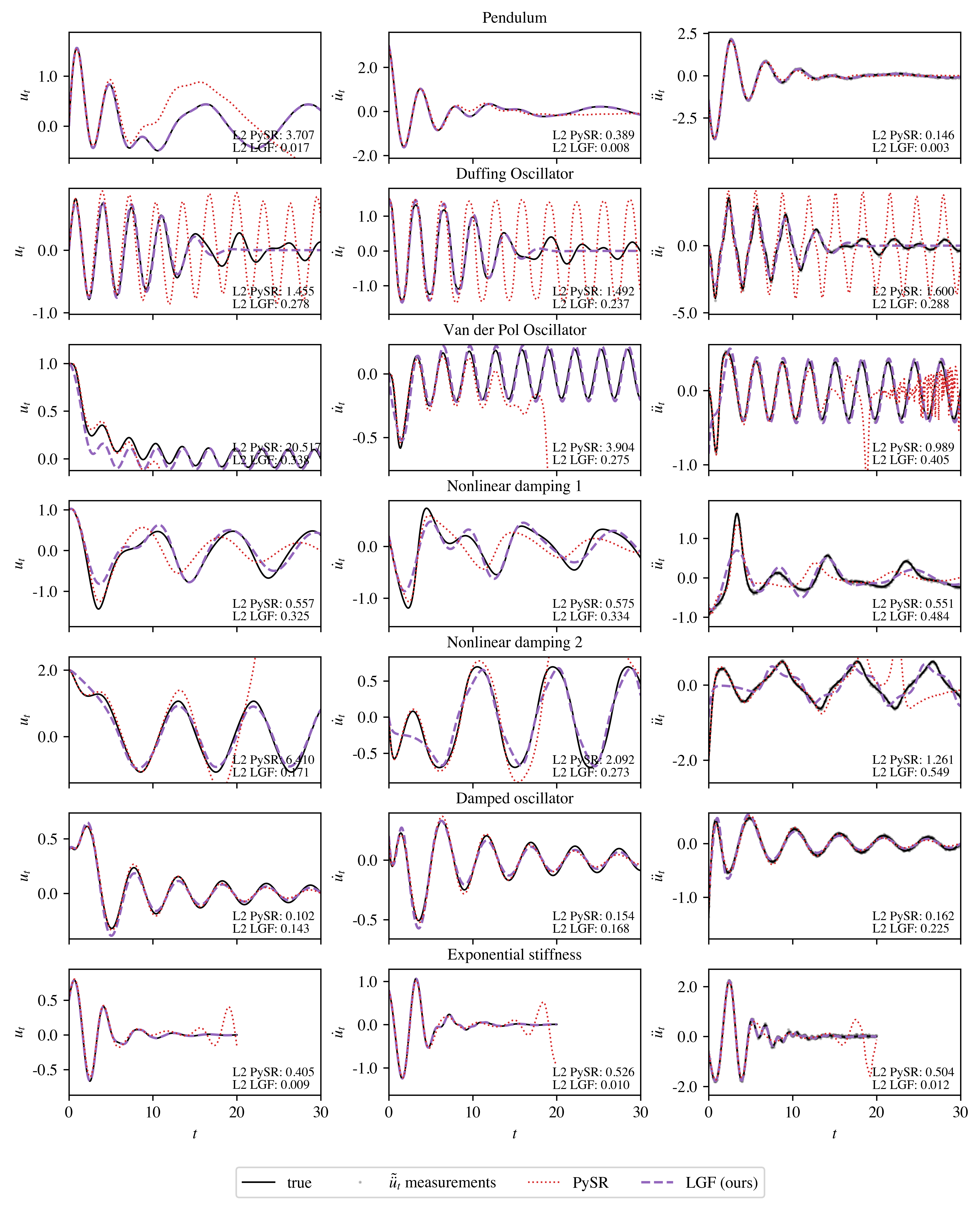}
    \caption{The numerical trajectories of the ground truth and predicted ODEs of Benchmark 3.}
    \label{fig:app_B3_num}
\end{figure}

The numerical trajectories of the displacement $u_t$, velocity $\dot{u}_t$ and acceleration $\ddot{u}_t$ of the ground truth and predicted ODEs are shown in \cref{fig:app_B3_num}.

%%%%%%%%%%%%%%%%%%%%%%%%%%%%%%%%%%%%%%%%%%%%%%%%%%%%%%%%%%%%%%%%%%%%%%%%%%%%%%%
%%%%%%%%%%%%%%%%%%%%%%%%%%%%%%%%%%%%%%%%%%%%%%%%%%%%%%%%%%%%%%%%%%%%%%%%%%%%%%%

\end{document}